# Products of Hidden Markov Models: It Takes N>1 to Tango

**Graham W. Taylor and Geoffrey E. Hinton**
Dept. of Computer Science
University of Toronto
Toronto, M5S 2Z9 Canada
{gwtaylor,hinton}@cs.toronto.edu

## Abstract

Products of Hidden Markov Models (PoHMMs) are an interesting class of generative models which have received little attention since their introduction. This may be in part due to their more computationally expensive gradient-based learning algorithm, and the intractability of computing the log likelihood of sequences under the model. In this paper, we demonstrate how the partition function can be estimated reliably via Annealed Importance Sampling. We perform experiments using contrastive divergence learning on rainfall data and data captured from pairs of people dancing. Our results suggest that advances in learning and evaluation for undirected graphical models and recent increases in available computing power make PoHMMs worth considering for complex time-series modeling tasks.

## 1 INTRODUCTION

Hidden Markov Models (HMMs) are statistical models of sequences that have proven successful in speech and language modeling (Rabiner, 1989; Bengio, 1999), and more recently, biological sequence analysis (Durbin et al., 1998). The advantage of an HMM over an $N$-th order Markov model is that it introduces a hidden state that controls the dependence of the current observation on the history of observations (Fig. 1a). However, many high-dimensional data sets with rich componential structure cannot be modeled efficiently by HMMs due to this simple but restrictive multinomial state. To model $K$ bits of information about the past history the HMM requires $2^K$ hidden states.

To avoid this exponential explosion we could condition the transition and emission distributions on another input sequence, such that these distributions change with time. Such a model is called the Input-Output Hidden Markov Model (IOHMM) (Bengio and Frasconi, 1995), also known as a non-homogeneous HMM (Fig. 1b). A more natural way, however, is to seek a model with distributed (i.e. componential) hidden state that has a representational capacity which is linear in the number of components. Linear dynamical systems satisfy this requirement, but they cannot model nonlinear dynamics.

Factorial Hidden Markov Models (FHMMs) (Ghahramani and Jordan, 1997) were introduced to address the need for distributed hidden state in HMMs. The FHMM generalizes the HMM by representing the state using a collection, instead of a single discrete state variable (Fig. 1c). However, FHMMs are directed models and thus observing a sequence introduces dependencies between the chains; an effect commonly known as "explaining away". This renders exact inference intractable and one must resort to approximate techniques such as Gibbs sampling or variational approximations to the posterior distribution. Products of Hidden Markov Models (Brown and Hinton, 2001), are a member of the Product of Experts family (Hinton, 2002), where each of the constituent experts is an HMM. Like the FHMM, the model provides a distributed state representation, but the relationship between hidden state and observations is now undirected (Fig. 1d). Conditioned on a sequence of observations, each HMM in the product is independent. Inference reduces to independently running the forward-backward algorithm in each HMM, and learning can be done efficiently by minimizing contrastive divergence (CD) (Hinton, 2002).

Since their introduction, PoHMMs have received little attention apart from some demonstrated success on "toy" language tasks (Brown and Hinton, 2001). They have yet to be applied to high-dimensional, real-world data. Part of the reluctance in adopting PoHMMs may have been due to the computationally demanding nature of gradient-based CD learning compared to the efficiency of the Baum-Welch algorithm for HMMs. However, since the introduction of PoHMMs, available computing power has greatly increased, and numerous insights have been made with respect to training models by CD (Tieleman, 2008). Another concern with undirected models in general has been the intractability of calculating the log likelihood of an observed sequence due to the existence of a normalizing factor (the "partition function") in the likelihood. In this paper, we demonstrate how the likelihood of sequences under a PoHMM can be estimated to a high degree of accuracy by estimating the partition function.

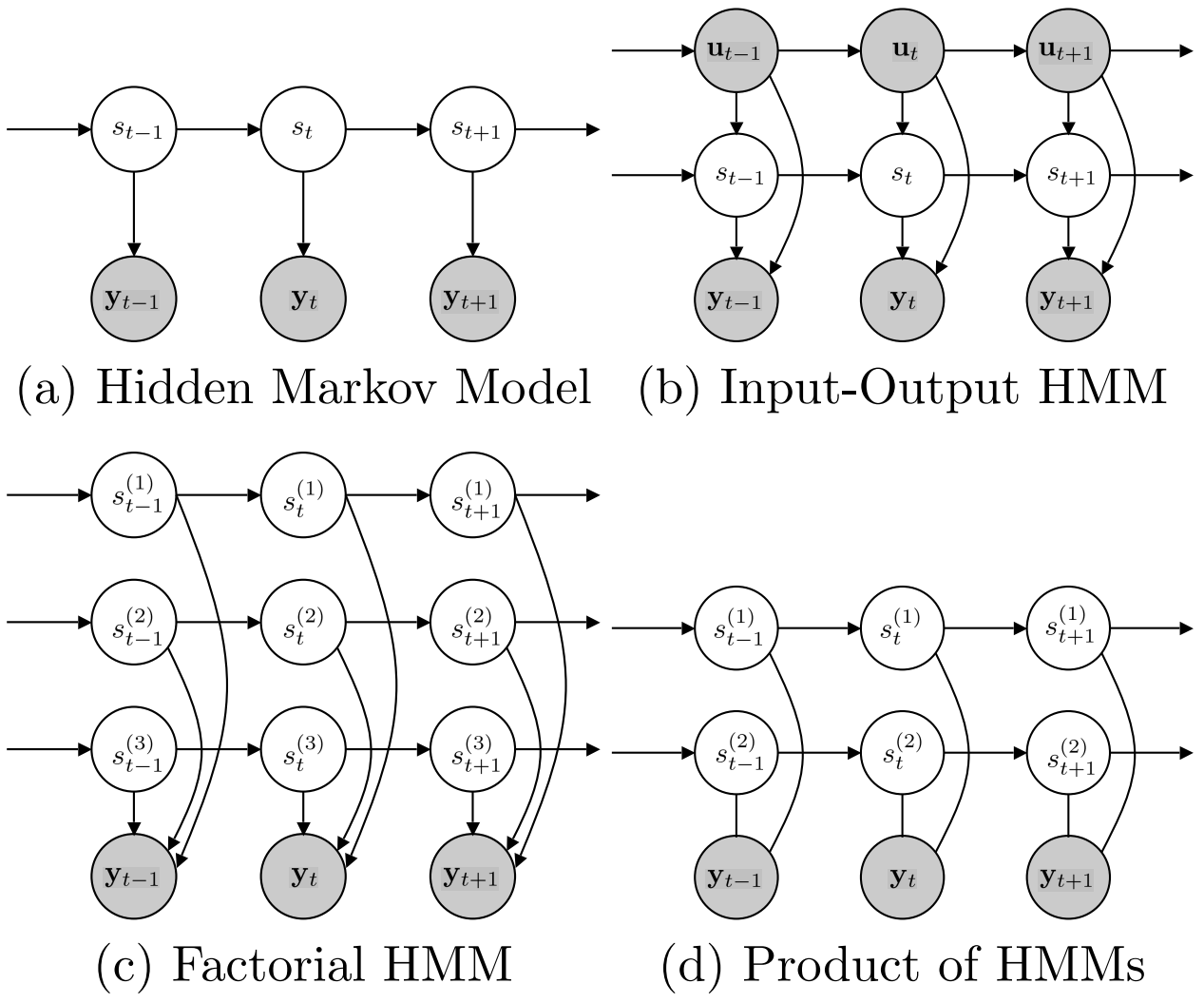

(a) Hidden Markov Model (b) Input-Output HMM (c) Factorial HMM (d) Product of HMMs

Figure 1: Graphical models specifying conditional independence relations for members of the HMM-based family.

## 2 PRODUCTS OF HIDDEN MARKOV MODELS

One way of modeling a complicated, high-dimensional data distribution is to combine a number of relatively simple models. Mixtures can approximate complicated smooth distributions arbitrarily accurately but are very inefficient in high dimensions. The mixture cannot produce a resultant distribution sharper than any of the individual experts. An alternative approach is to multiply the individual probability distributions together and then renormalize. This allows us to produce much sharper distributions than the individual models. The framework is called a Product of Experts (PoE) (Hinton, 2002). PoEs are defined by the following likelihood over observations:

$$p(\mathbf{y}|\Theta) = \frac{\prod_{m=1}^{M} p^{(m)}(\mathbf{y}|\theta^{(m)})}{\sum_{\mathbf{y}'} \prod_{m=1}^{M} p^{(m)}(\mathbf{y}'|\theta^{(m)})} \tag{1}$$

where $m$ indexes individual experts, $\Theta = \{\theta^{(m)}\}$ represents parameters, and the probability distribution of the individual experts may be of any type (including unnormalized potential functions). Learning a PoE appears to be difficult due to the partition function that appears in the expression for the likelihood and consequently the gradients for maximum likelihood learning. While exact learning is intractable, one can approximately minimize an alternative loss function which, in practice, works very well (provided the individual experts themselves are tractable). This method is known as contrastive divergence.

If we take each of the experts to be an HMM, the resulting product model is called a Product of Hidden Markov Models (PoHMM) (Brown and Hinton, 2001). Note that each expert now defines a distribution over sequences, $\{\mathbf{y}_t\}$, rather than static observations. PoHMMs are ideal for modeling data that is caused by multiple underlying influences. Each expert can model some aspect of the data and constrain it such that the combination of the models will have a very sharp distribution. Specifically, if each HMM can remember a different piece of information about the past, the PoHMM should be able to capture more long-range structure than a standard HMM.

CD learning updates the parameters according to:

$$\begin{aligned}\Delta\theta^{(m)} \propto & \left\langle \frac{\partial}{\partial\theta^{(m)}} \log p(\{\mathbf{y}_t\}|\theta^{(m)}) \right\rangle_{P_0} \\ & - \left\langle \frac{\partial}{\partial\theta^{(m)}} \log p(\{\mathbf{y}_t\}|\theta^{(m)}) \right\rangle_{P_K}\end{aligned} \tag{2}$$

where the first expectation is with respect to the data distribution, and the second expectation is with respect to the distribution of samples obtained after $K$ steps of alternating Gibbs sampling, initialized at the data. Typically $K = 1$, but recent results show that gradually increasing $K$ with learning can significantly improve performance at a modest additional computational cost (Carreira-Perpinan and Hinton, 2005). We refer to this technique as CD($K$). The gradient terms in Eq. 2 are obtained by the following procedure:

1. Set $\{\mathbf{y}_t\}^0$ to the observed sequence. Calculate each HMM's gradient, $\frac{\partial}{\partial\theta^{(m)}} \log p(\{\mathbf{y}_t\}^0|\theta^{(m)})$, on this sequence, using the forward-backward (f-b) algorithm. This leads to the first term in Eq. 2.
2. **Repeat for** $k = 1, \ldots, K$**:** Sample from the posterior distribution of paths through state space for each of the HMMs given $\{\mathbf{y}_t\}^{k-1}$ (we perform f-b independently in each HMM). Multiply together the emission distributions specified by the sampled paths in each HMM and renormalize. Draw a new visible sequence, $\{\mathbf{y}_t\}^k$, from this distribution.

3. For each HMM, calculate $\frac{\partial}{\partial \theta^{(m)}} \log p(\{\mathbf{y}_t\}^K|\theta^{(m)})$ on the final $K$-step reconstructed sequence, using f-b. This leads to the second term in Eq. 2.

The learning rule is not a true gradient, so we cannot employ more sophisticated second-order methods or conjugate gradient optimization. The convergence time will be much slower than for fixed-point methods such as the EM algorithm used to fit tractable models like the standard HMM. Despite this shortcoming, the PoHMM shows some clear advantages to standard HMMs both in theory and in practice.

# 3 ESTIMATING THE PARTITION FUNCTION

Following Eq.1 the log likelihood that a PoHMM assigns to a sequence,$\{\mathbf{y}_t\}$, is given by:

$$\log p(\{\mathbf{y}_t\}|\Theta) = \sum_{m=1}^{M} \log p^{(m)}(\{\mathbf{y}_t\}|\theta^{(m)}) - \log Z \quad (3)$$

where $Z = Z(\Theta, T)$ is the partition function which depends on the model parameters and length of the sequence. A key feature of the PoHMM is the fact that given an observation, the experts decouple and the first term can be easily computed by evaluating $\log p^{(m)}(\{\mathbf{y}_t\}|\theta^{(m)})$ for each expert. This can be found by summing out the hidden variables from the joint likelihood, which is done efficiently in an HMM via the forward-backward algorithm. The partition function, however, involves an intractable sum (or integral) over all possible observations of length $T$. For all but the smallest discrete models, it is not practical to compute.

A good estimate of the partition function would aid us in both model selection and controlling model complexity since it allows us to compute an estimate of the log probability of an observation under the model. It would also permit the comparison of PoHMMs to other generative models, such as standard HMMs. Salakhutdinov and Murray (2008) have recently reported success in applying a particular type of Monte Carlo method, Annealed Importance Sampling (AIS), to another type of Product of Experts, the Restricted Boltzmann Machine (RBM). Given the similarity between members of the PoE family, we are motivated to extend their work to PoHMMs. We first give a brief review of AIS but refer the reader to Neal (2001) for a more thorough discussion. We will assume the observable space is over discrete sequences, $\{\mathbf{y}_t\}$, but the development can readily be extended to continuous PoHMMs (e.g. Gaussian emission distributions).

## 3.1 Importance sampling

Importance sampling is a Monte Carlo method that allows us to either generate samples from a complicated distribution of interest, or calculate statistics with respect to that distribution. For the purposes of our discussion, we call this distribution $p_B$. Importance sampling is based on the idea of generating independent points from some simpler approximating distribution, $p_A$, and associating a weight with each point to compensate for the use of the wrong distribution. A byproduct of importance sampling is that it also provides an estimate of the ratio of the partition functions of the two distributions. This ratio is a useful quantity because it allows us to compare two different models by the likelihood they assign to some test data. To see this, suppose that $p_A(\{\mathbf{y}_t\}|\theta_A) = \frac{p_A^*(\{\mathbf{y}_t\}|\theta_A)}{Z_A(\theta_A,T)}$ and $p_B(\{\mathbf{y}_t\}|\theta_B) = \frac{p_B^*(\{\mathbf{y}_t\}|\theta_B)}{Z_B(\theta_B,T)}$ are sequence models (e.g. PoHMMs) where $p_A^*$ and $p_B^*$ are computable but $Z_A$ and $Z_B$ are not. Comparing likelihoods leads to:

$$\frac{p_A(\{\mathbf{y}_t\}|\theta_A)}{p_B(\{\mathbf{y}_t\}|\theta_B)} = \frac{p_A^*(\{\mathbf{y}_t\})}{p_B^*(\{\mathbf{y}_t\})}\frac{Z_B}{Z_A} \quad (4)$$

where we use the shorthand notation introduced in the rightmost term to simplify the presentation. Eq. 4 states that to compare the models based on likelihood, we must be able to compute the ratio of partition functions, $Z_B/Z_A$. Note that we can rewrite the ratio as:

$$\frac{Z_B}{Z_A} = \frac{\sum_{\{\mathbf{y}_t\}} p_B^*(\{\mathbf{y}_t\})}{Z_A} = \sum_{\{\mathbf{y}_t\}} \frac{p_B^*(\{\mathbf{y}_t\})}{p_A^*(\{\mathbf{y}_t\})} p_A(\{\mathbf{y}_t\})$$
$$= \mathbb{E}_{p_A}\left[\frac{p_B^*(\{\mathbf{y}_t\})}{p_A^*(\{\mathbf{y}_t\})}\right] \quad (5)$$

assuming that $p_A^*(\{\mathbf{y}_t\}) \neq 0$ whenever $p_B^*(\{\mathbf{y}_t\}) \neq 0$. The sum (or integral) is taken over all sequences of length $T$.

If we can draw independent samples from $p_A$, importance sampling allows us to estimate Eq. 5 by:

$$\frac{Z_B}{Z_A} \approx \frac{1}{P}\sum_{p=1}^{P} \frac{p_B^*(\{\mathbf{y}_t\}^{(p)})}{p_A^*(\{\mathbf{y}_t\}^{(p)})} \equiv \frac{1}{P}\sum_{p=1}^{P} w^{(p)} \quad (6)$$

where $\{\mathbf{y}_t\}^{(p)} \sim p_A$ and $w^{(p)}$ is the *importance weight* of sample $p$. However, if $p_A$ and $p_B$ differ considerably then Eq. 6 is a poor estimate of the true ratio $Z_B/Z_A$.

## 3.2 Annealed importance sampling

Annealed importance sampling (Neal, 2001) can achieve better estimates of $Z_B/Z_A$ by defining a series of intermediate distributions, $p_0, p_1, \ldots, p_N$ which slowly anneal from $p_0 = p_A$ to $p_N = p_B$. The intermediate distributions must be defined such that we can easily evaluate the unnormalized probability, $p_n^*(\{\mathbf{y}_t\}); \forall\{\mathbf{y}_t\}, n = 0, 1, \ldots, N$. We also must define a Markov Chain transition operator, $T_n(\{\mathbf{y}_t\}, \{\mathbf{y}_t\}')$ that leaves $p_n(\{\mathbf{y}_t\})$ invariant:

$$\int T_n(\{\mathbf{y}_t\}, \{\mathbf{y}_t\}')p_n(\{\mathbf{y}_t\})d\{\mathbf{y}_t\} = p_n(\{\mathbf{y}_t\}'). \quad (7)$$

AIS proceeds by a series of $P$ runs, one per sample. Given the definition of $p_n(\{\mathbf{y}_t\})$ and $T_n(\{\mathbf{y}_t\}, \{\mathbf{y}_t\}')$, a single AIS run is carried out as follows:

- Sample $\{\mathbf{y}_t\}^0$ from $p_0 = p_A$
- **Repeat for** $n = 1, 2, \ldots, N-1$:
  Sample $\{\mathbf{y}_t\}^n$ from $\{\mathbf{y}_t\}^{n-1}$ using $T_{n-1}$
- Set $w^{(p)} = \frac{p_1^*(\{\mathbf{y}_t\}^0)}{p_0^*(\{\mathbf{y}_t\}^0)} \frac{p_2^*(\{\mathbf{y}_t\}^1)}{p_1^*(\{\mathbf{y}_t\}^1)} \cdots \frac{p_N^*(\{\mathbf{y}_t\}^{N-1})}{p_{N-1}^*(\{\mathbf{y}_t\}^{N-1})}$

After performing $P$ runs of AIS we can estimate $Z_B/Z_A$ using Eq. 6. The estimate is unbiased. Its variance depends on the number of runs, $P$, and the number of intermediate distributions, $N$ (see Neal, 2001).

### 3.3 AIS for PoHMMs

AIS for any particular model depends on the form of the intermediate distributions, $p_n(\{\mathbf{y}_t\})$, and the Markov chain transition operator, $T_n(\{\mathbf{y}_t\}, \{\mathbf{y}_t\}')$. A general form for the intermediate distributions is:

$$p_n(\{\mathbf{y}_t\}) \propto p_A^*(\{\mathbf{y}_t\})^{(1-\beta_n)} p_B^*(\{\mathbf{y}_t\})^{\beta_n} \quad (8)$$

where $0 = \beta_0 < \beta_1 < \ldots < \beta_N = 1$ is the user-specified annealing schedule. If $p_A$ and $p_B$ are defined by PoHMMs, this leads to an unnormalized distribution for which both evaluating $p_n^*$ and sampling are not straightforward. However, we can use the following distribution:

$$\begin{aligned} p_n(\{\mathbf{y}_t\}) &= \frac{p_n^*(\{\mathbf{y}_t\})}{Z_n(\Theta, T)} \\ &= \Bigg[\prod_{m=1}^{M_A} \sum_{\{s_t^{(m)}\}} p^{(m)}(s_1^{(m)}) \prod_{t=2}^{T} p^{(m)}(s_t^{(m)}|s_{t-1}^{(m)}) \\ &\times \prod_{t=1}^{T} p^{(m)}(\mathbf{y}_t|s_t^{(m)})^{(1-\beta_n)}\Bigg] \Bigg[\prod_{m=1}^{M_B} \sum_{\{s_t^{(m)}\}} p^{(m)}(s_1^{(m)}) \\ &\times \prod_{t=2}^{T} p^{(m)}(s_t^{(m)}|s_{t-1}^{(m)}) \prod_{t=1}^{T} p^{(m)}(\mathbf{y}_t|s_t^{(m)})^{\beta_n}\Bigg] \Big/ Z_n \quad (9) \end{aligned}$$

where $m, n$ index the $M_A, M_B$ HMMs in PoHMM $A, B$, respectively. Eq. 9 defines a PoHMM with $M_A + M_B$ chains, one group whose emission distributions are scaled by $1 - \beta_n$ and the other group whose emission distributions are scaled by $\beta_n$. When $\beta_n = 0$, $p_n$ reduces to PoHMM $A$ with $M_A$ chains, and when $\beta_n = 1$, $p_n$ reduces to PoHMM $B$ with $M_B$ chains.

Similar to (Salakhutdinov and Murray, 2008), $T_n(\{\mathbf{y}_t\}, \{\mathbf{y}_t\}')$ is defined by alternating Gibbs sampling between $\{\mathbf{y}_t\}^n$ and the $M_A + M_B$ hidden states:

- Given the current sample, $\{\mathbf{y}_t\}$, perform inference in each of the $M_A$ HMMs in PoHMM $A$ using the f-b algorithm with the scaled emission distribution, $p^{(m)}(\mathbf{y}_t|s_t^{(m)})^{(1-\beta_n)}$. Sample a hidden state sequence, $\{s_t^{(m)}\}$, from each posterior.
- Repeat the above step for each of the $M_B$ HMMs in PoHMM $B$, but using $p^{(m)}(\mathbf{y}_t|s_t^{(m)})^{\beta_n}$.
- Conditional on $\{s_t^{(m)}\}\forall m \in M_A$ and $\{s_t^{(m)}\}\forall m \in M_B$, the contribution to the distribution over observables from each of the HMMs will be $p^{(m)}(\mathbf{y}_t|s_t^{(m)})^{(1-\beta_n)}$ and $p^{(m)}(\mathbf{y}_t|s_t^{(m)})^{\beta_n}$, respectively. For each $t$, multiply the $(M_A + M_B)$ conditional distributions together and renormalize.
- Draw $\{\mathbf{y}_t\}'$ from the normalized conditional distribution over observables.

Our definition of $T_n(\{\mathbf{y}_t\}, \{\mathbf{y}_t\}')$ leaves $p_n$ invariant. Since $p_n$ defines a PoHMM, $p_n^*$ is easily evaluated via the forward-backward algorithm using the scaled emission probability distributions, as in inference. The resampling and likelihood estimation steps necessary for AIS can be combined into a single forward-backward pass (per annealing step) in each HMM. Although we can obtain samples from $p_A$ by alternating Gibbs sampling, they will not be independent. It is preferable to use independent samples, but AIS will still converge to the correct estimate, provided that the Markov chain is ergodic (Neal, 2001).

If we wish to calculate $Z_B$ instead of $Z_B/Z_A$, we can select PoHMM $A$ to have a $Z_A$ that is easily evaluated. For example, if $M_A = 1$ then PoHMM $A$ is just a standard HMM and $Z_A = 1$. In practice, we use a single-state HMM whose emission distribution is estimated to be the base rates of the training data, smoothed so that $p_A(\cdot)$ is never zero. We choose the $\beta_k$ such that we gently anneal the base-rate HMM to the PoHMM of interest. This is why in Eq. 9 we do not need to scale the terms related to the dynamics: the dynamics are completely defined by the PoHMM.

## 4 MODELING DAILY RAINFALL OCCURRENCE

Precipitation modeling is of interest to fields such as hydrology, geophysics, and agriculture. Existing models are challenged by the large variance of observa-tions over time, as well as the local effects present in the data. Recently, HMMs have been proposed as a way to describe daily rainfall occurrence data collected at several weather stations (Kirshner, 2005), as they are able to capture both spatial dependencies between weather stations and temporal regularities. The hidden states of the HMMs can also be interpreted as representing "wet-dry" or directional effects and the state sequences can be analyzed for the existence of seasonal, intraseasonal, interannual and longer-scale time patterns. Furthermore, the generative nature of the HMM allows for the production of station-scale daily rainfall simulations for input into other systems such as crop models.

For comparison purposes, we will consider one of the rainfall data sets which has already been extensively analyzed in the context of HMMs. The Ceará data set contains 24 sequences of 90 binary observations for each of 10 weather stations located in northeast Brazil. The sequences correspond to the wet seasons from the years 1975-2002, however, four of the years (1976, 1978, 1984, 1986) contained a significant number of missing observations and were removed. All seasons start February 1 and end in April. As a baseline model, we consider a simple HMM with Bernoulli emission distributions on each dimension that are conditionally independent, given the state (denoted HMM-CI). Thus, the spatial dependencies between weather stations are captured entirely by the hidden state, and the emission parameters for each dimension can be estimated independently. Kirshner (2005) has looked at two ways of extending this HMM baseline. The first is to employ an IOHMM, for example, conditioning the transition distributions of the HMM on the output of a coarser-resolution General Circulation Model that predicts average seasonal precipitation. The second is to experiment with emission distributions with richer dependency structure which can capture interactions not only between weather stations on a given day, but between neighbouring observations. In both cases, the multinomial hidden state is left intact. We propose maintaining the simple conditionally-independent Bernoulli emission distribution but using a distributed hidden state.

We compare baseline HMMs of $\{2,\ldots,8\}$ states to a product of two HMMs of $\{2,\ldots,6\}$ states and a product of three HMMs of $\{2,\ldots,6\}$ states. We use the same notation as Kirshner (2005) for the various HMMs. In addition to HMM-CI (as noted above), we experimented with a first-order Autoregressive HMM with dependence on the previous observation for the same station (HMM-Chains), an HMM with Chow-Liu tree emissions (HMM-CL), an HMM with conditional Chow-Liu forest emissions (HMM-CCL) and an HMM whose output distribution is a full bivariate maximum entropy model (HMM-MaxEnt).

Parameters are initialized randomly. The HMMs are trained with the EM algorithm until convergence is observed in the training log likelihood. The PoHMMs are trained with CD(1) for a fixed number of epochs at a conservative learning rate (0.001 for all parameters). Following (Kirshner, 2005), we use *leave-6-out* cross-validation. Under *leave-6-out* we train 4 models, one for each set obtained by leaving out 6 non-overlapping consecutive sequences. Each model is then evaluated on the corresponding left-out 6 sequences. We report the mean of the 4 models. This entire process is repeated 100 times (using different random initializations) and we report the mean of these runs. We employed four different evaluation metrics:

**Scaled log likelihood** For the HMM, the log likelihoods of the held out sequences are computed exactly using the forward-backward algorithm. For the PoHMMs, we use 100 AIS runs with 500 intermediate distributions of uniform spacing[1] to estimate the partition function. Log likelihoods of the held out sets are divided by the number of binary events occurring in the held out data ($6 \times 90 \times 10$).

**Classification accuracy** We report the average classification accuracy in predicting observations that are removed (one dimension, one sequence at a time) from the held out data. Since observations are binary, we simply compare the (unnormalized) log likelihood of the sequence with the correct observation filled in vs. the incorrect observation.

**Difference in precipitation persistence for observed and simulated data** Persistence is defined as the probability of a precipitation occurrence for a particular station given a precipitation event at that station at the previous observation. We simply compared the mean of this result over 500 simulated 90-day sequences to the mean calculated over the held-out observations and report the absolute difference.

**Difference in correlation for observed and simulated data** The spatial correlation between a pair of stations is computed as Pearson's coefficient of their respective daily rainfall occurrences. We compute this value separately for the held-out sequences and the simulated data and report the absolute difference.

The results are shown in Fig. 2. Given a fixed number of parameters, PoHMMs outperform the various HMMs under the log likelihood and classification metrics. In the case of log likelihood, the PoHMMs also exhibit less overfitting of the training data (though this could also be an artifact of a fixed number of epochs vs. training until convergence). They also do well on the persistence task, though to a lesser degree. Not surprisingly, the best performing model according to the difference in persistence is HMM-Chains which has stronger temporal structure. PoHMMs do not perform as well according to the difference in spatial correlation. Perhaps this is to be expected as they retain the same simple emission model as HMM-CI. Models with stronger spatial structure (HMM-CL,HMM-MaxEnt) perform better under this metric. Another interesting observation is that if we examine the emission parameters of the PoHMMs, we can see that individual HMMs specialize locally (Fig. 3).

[1] A nonlinear (e.g. logarithmic) schedule may give better results, but quick initial runs suggested that it was not necessary. It may also require additional tunable parameters.

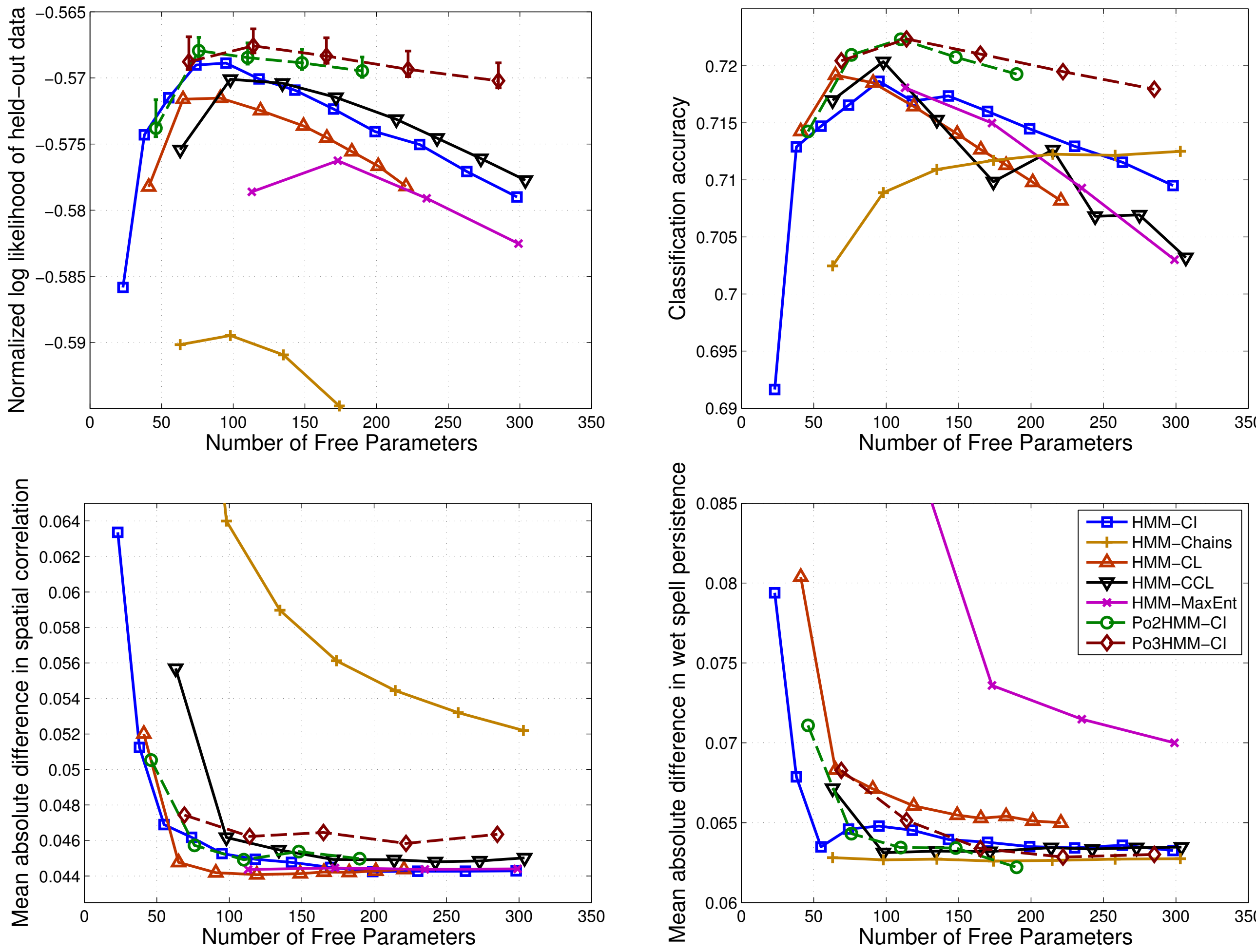

Figure 2: Rainfall modeling. We have compared several types of HMMs (solid lines), to a product of two or three HMMs (dashed lines) under a variety of metrics. In the top two plots, higher is better. In the bottom two plots, lower is better. Log likelihoods are per-sequence, per-frame, per-dimension. Error bars of three standard deviations of the partition function estimate are plotted for log likelihood. The single legend applies to all plots.

## 5 MODELING HUMAN DANCE

Data captured from human motion (mocap) is often high-dimensional and contains complex nonlinear relationships between variates. This makes learning challenging for simple models. Generative models with distributed hidden state have had success modeling locomotion (Taylor et al., 2007), however, this type of motion exhibits strong global coherence and should not pose as great a difficulty to an HMM as it can allocate subsets of its states to the various regimes. At each frame, it is not difficult to predict what one part of the body is doing from another. The advantages of componential state should be more apparent in modeling data that is the result of multiple underlying processes, such as mocap from multiple subjects.

We carried out a series of experiments to compare HMMs and PoHMMs on data with both weakly and strongly componential structure. We started with mocap data of a pair of people salsa dancing (all frames from subjects 60 and 61 in the CMU motion capture database). To facilitate training and comparison, the data was partitioned into fixed-length sequences of 100 frames. The ordering of sequences was randomly permuted so that training and test sets were balanced and randomly assigned. 23 sequences were set aside for testing and the remaining 48 sequences were used for training. We used the same representation as described in (Taylor et al., 2007) where all joint angles were encoded via the exponential map and a local coordinate system was employed to ensure invariance to translations in the ground plane and rotations about the vertical. In order to work with discrete HMMs while still maintaining the multivariate and componential properties of the mocap data, we performed a type of vector quantization. Within each subject, dimensions were assigned to one of 6 local groups: Left leg, Right leg, Torso, Upper body, Left arm and Right arm. ($K = 25$)-means clustering was applied independently to the vectors corresponding to each group, across all frames. This resulted in 6 or 12-dimensional discrete sequences for one or two subjects, respectively.

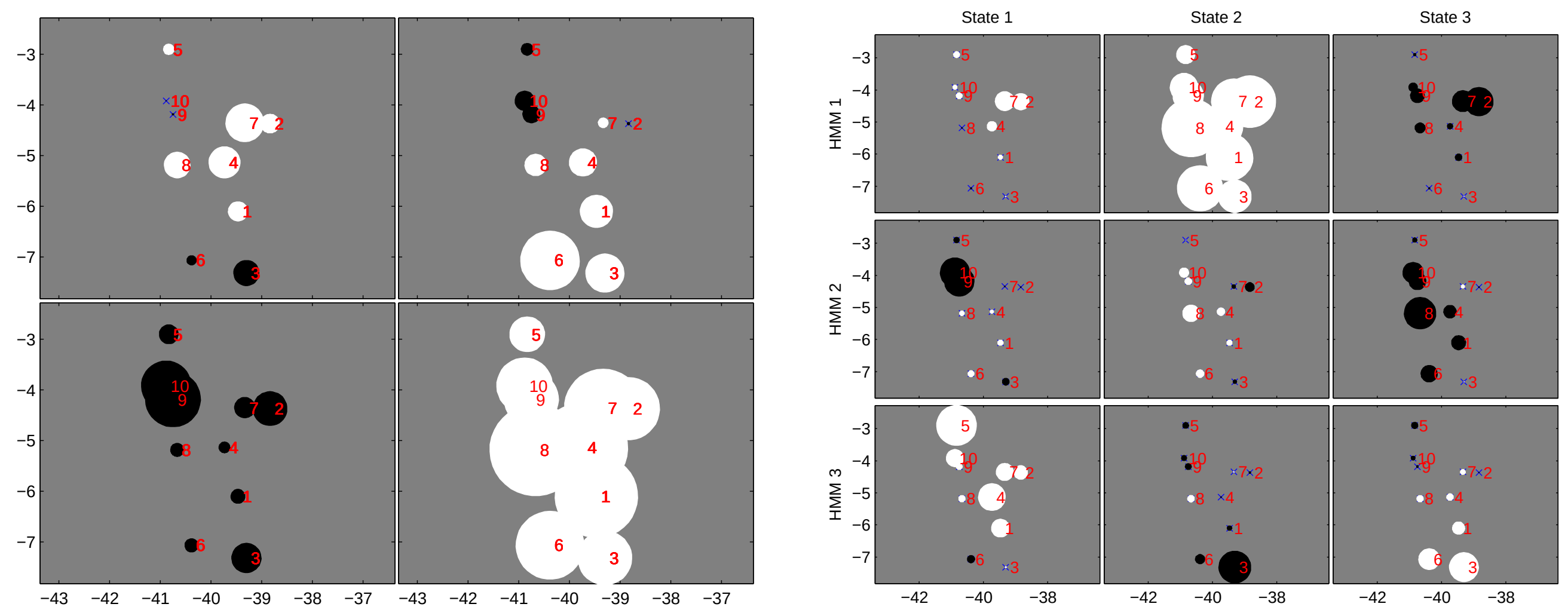

Figure 3: The log-domain emission parameters of a four-state HMM (left) and product of three three-state HMMs (right). Each rectangle reflects the spatial layout of the numbered weather stations which correspond to input dimensions. White represents a belief in a dry observation while black represents a belief in a wet observation, conditional on that state. The larger the radius, the stronger the belief. In the product, HMM 1 captures eastern effects, HMM 2 captures western effects, and HMM 3 captures northern and southern effects.

All emission distributions were 6 or 12 25-symbol multinomials conditionally independent given hidden state. Baseline HMMs of $2^N$ states where $N = 2, \ldots, 6$ were trained with EM until convergence based on training log likelihood. We also trained several PoHMM models, where the number of experts and states were selected such that the number of free parameters were comparable to the baseline HMMs. All PoHMMs were trained using CD(10) which gave slightly better results than our initial models obtained using CD(1). The PoHMMs were all trained for 1000 epochs over all sequences, using a learning rate of 0.01 for all parameters. A momentum term was also used: 0.9 of the previous accumulated gradient was added to the current gradient.

Log likelihoods under the HMMs were computed exactly. To estimate log likelihoods under the PoHMMs, we used 100 AIS runs. We used an annealing schedule of 5000 uniformly spaced intervals. Figure 4 shows the results obtained modeling single-subject mocap data (weakly componential) and two-subject mocap data (strongly componential). All log likelihoods are per sequence, and have been normalized by dividing by the number of dimensions (6 and 12 respectively) so that the two plots are comparable.

In both cases, products of HMMs compare favourably to the baseline. As expected, the advantage is more evident when the data has strongly componential structure. This is achieved despite a weaker gradient-based method of training. We note that HMMs trained by conjugate-gradient or CD(1) gave worse baseline results than EM. All of these methods rely on inference via the forward-backward algorithm whose running time is quadratic in the number of hidden states. This suggests that combining HMMs with a smaller number of states via a product where training time is linear in the number of HMMs is a sensible thing to do. However, the cost of gradient-based learning compared to EM means that this advantage will not be seen until the number of states are considerably higher than those considered in our experiments. All of the PoHMMs we trained took on the order of an hour, while the HMMs were trained within minutes.

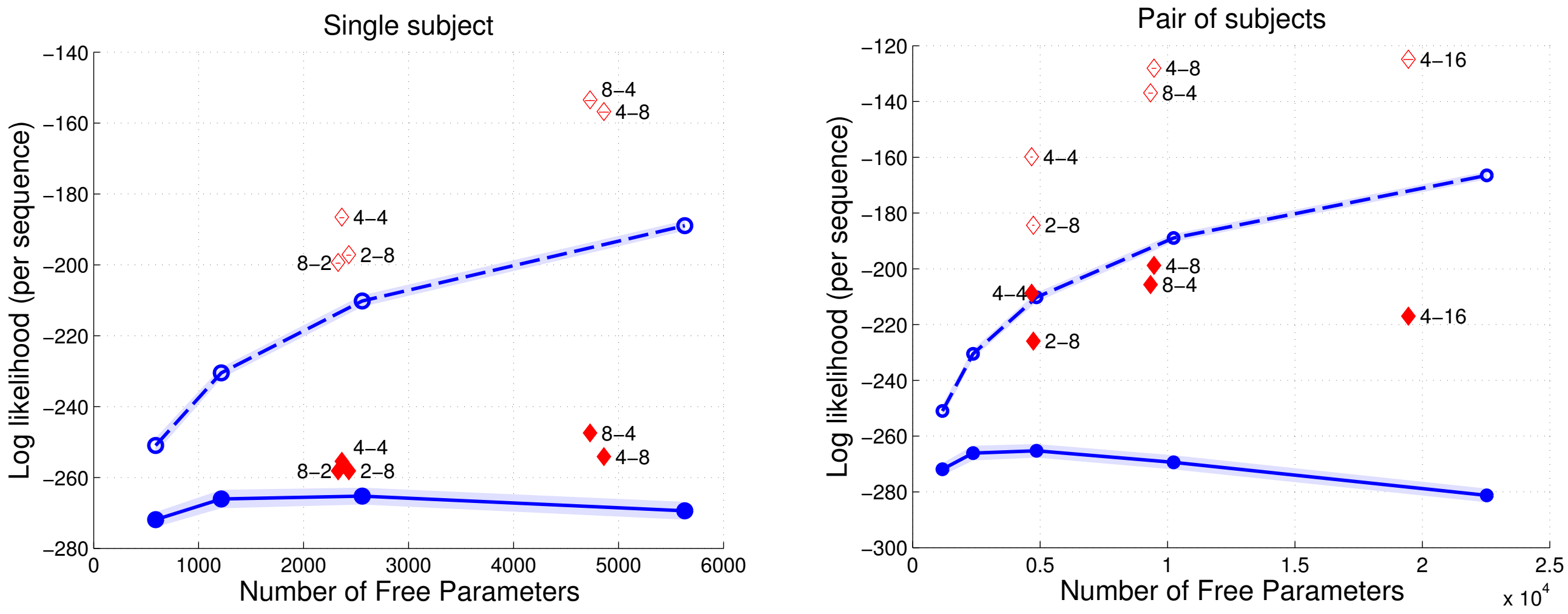

Figure 4: Modeling a single subject (left) and partners (right) salsa dancing. Open and solid circles correspond to the log likelihood scores of the baseline HMMs on the training and test data. Open and solid diamonds denote training and test log likelihood scores for the PoHMMs. The small text beside the PoHMM scores indicates the number of HMMs and number of states per HMM in the product. Error bars corresponding to three standard deviations from the partition function estimate have been plotted (visible in open diamonds) but they are so tight that they are difficult to see at this scale. Shading indicates unit standard deviation error across HMMs.

## 6 DISCUSSION

Advances in computing hardware and improvements to the CD learning algorithm have made Products of Hidden Markov Models a more attractive option for time-series modeling than when they were first introduced. This is especially true for data that is high-dimensional and has strongly componential structure. With a reliable means of approximating the intractable partition function, log likelihood estimates of sequences can aid in model selection, complexity control and comparisons to other generative models.

In this work, we have departed from the domain of "toy" problems and demonstrated the effectiveness of PoHMMs on real multivariate data. In precipitation modeling, PoHMMs were shown to improve performance over HMMs of a similar number of parameters under a variety of metrics and also captured interesting local regularities in the data. When trained on data captured from human motion, PoHMMs were also shown to compare favourably to HMMs in terms of log likelihood. This was most apparent when the data was a product of multiple underlying influences.